\documentclass[conference]{IEEEtran}
\IEEEoverridecommandlockouts
\usepackage{makecell}
\usepackage{amsmath} % assumes amsmath package installed
\usepackage{amssymb}  % assumes amsmath package installed

\usepackage{amsfonts}
\usepackage{amssymb}
\usepackage{amsmath}
\usepackage{graphics} % for pdf, bitmapped graphics files
\usepackage{epsfig} % for postscript graphics files
\usepackage{mathptmx} % assumes new font selection scheme installed
\usepackage{times} % assumes new font selection scheme installed
\usepackage{subfigure}

\usepackage{mathrsfs}
\usepackage{indentfirst}
\usepackage{multirow}

\usepackage{cite}

\usepackage{algorithmic}
\usepackage{graphicx}
\usepackage{textcomp}
\usepackage{xcolor}
\usepackage{bbding}
\usepackage{geometry}
\usepackage[colorlinks=true,linkcolor=blue,citecolor=green,urlcolor=blue,]{hyperref}

\def\BibTeX{{\rm B\kern-.05em{\sc i\kern-.025em b}\kern-.08em
    T\kern-.1667em\lower.7ex\hbox{E}\kern-.125emX}}
\begin{document}

\title{	Ground-Challenge: A Multi-sensor SLAM Dataset	Focusing on Corner Cases for Ground Robots

\author{Jie Yin $^{\dag}$, Hao Yin $^{\dag}$, Conghui Liang$^{*}$ $^{\ddag}$ and Zhengyou Zhang $^{\ddag}$ (IEEE Fellow $\&$ ACM Fellow)
% \thanks{Manuscript received: September 8th, 2021; Revised:
% December 2rd, 2021; Accepted: December 12th, 2021.}%
% \thanks{This paper was recommended for publication by
% Editor Javier Civera upon evaluation of the Associate Editor and Reviewers' comments.}%

\thanks{ Authors $^{\dag}$ are independent researchers. Authors $^{\ddag}$  are with Tencent Robotics X Lab, Shenzhen, China.
$^*$ Corresponding Author: Conghui Liang ({\tt\small isaacliang@tencent.com})}}

}

\maketitle

\begin{abstract}
High-quality datasets can speed up breakthroughs and reveal potential developing directions in SLAM research.
To support the research on corner cases of visual SLAM systems,
this paper presents Ground-Challenge: a challenging dataset comprising 36 trajectories with diverse corner cases such as aggressive motion, severe occlusion, changing illumination, few textures, pure rotation, motion blur, wheel suspension, etc. The dataset was 
collected by a ground robot with multiple sensors including an RGB-D camera, an inertial measurement unit (IMU), a wheel odometer and a 3D LiDAR. All of these sensors were well-calibrated and synchronized, and their data were recorded simultaneously.
To evaluate the performance of cutting-edge SLAM systems, we tested them on our dataset and demonstrated that these systems are prone to drift and fail on specific sequences.
We will release the full dataset and relevant materials upon paper publication to benefit the research community. For more information, visit our project website at \href{https://github.com/sjtuyinjie/Ground-Challenge}{https://github.com/sjtuyinjie/Ground-Challenge}.
\end{abstract}


	\begin{figure*}
		\begin{center}
			\begin{tabular}{cccc}
				\includegraphics[scale=0.024]{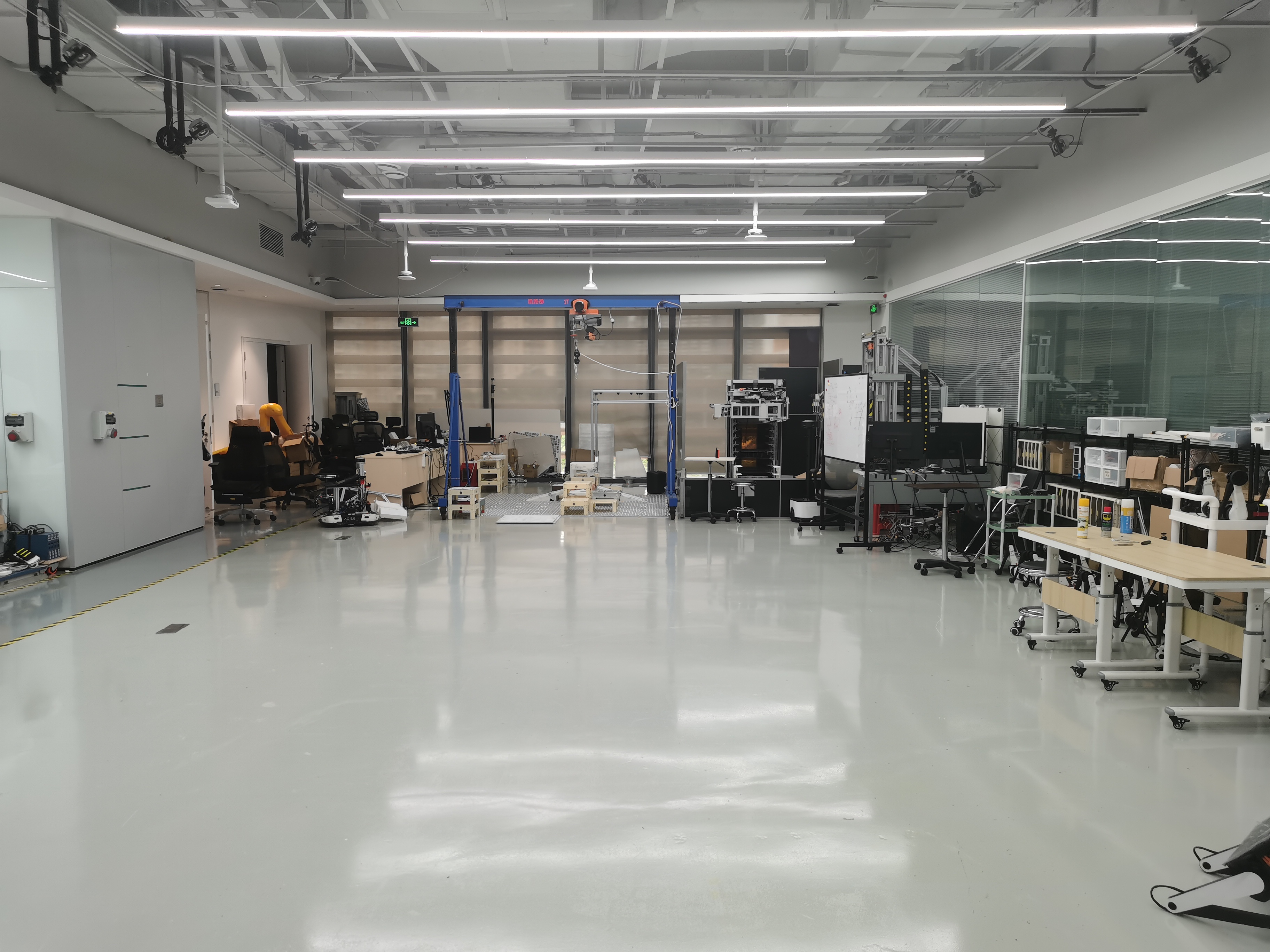} &
				\includegraphics[scale=0.024]{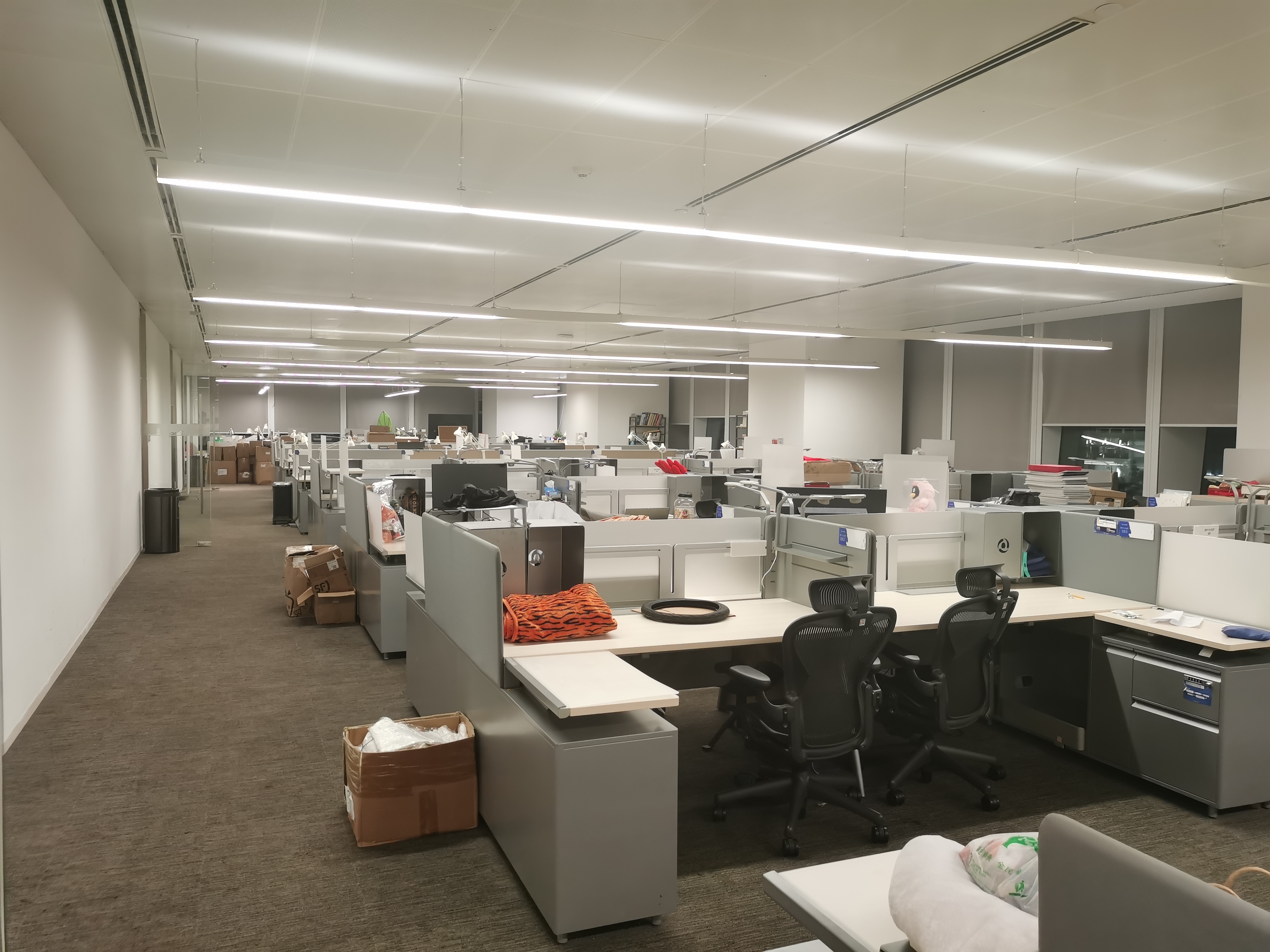} &
				\includegraphics[scale=0.024]{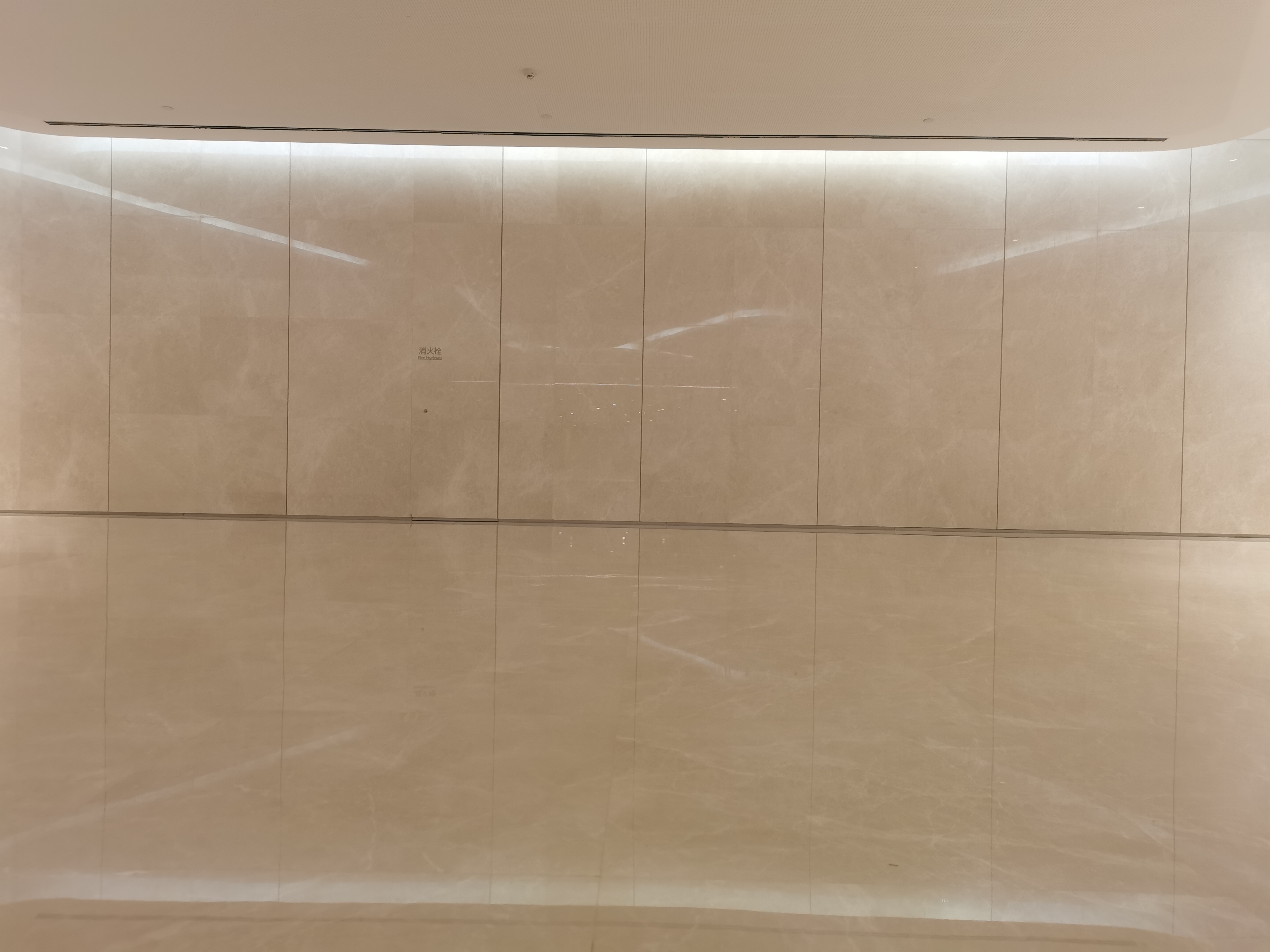}&
				\includegraphics[scale=0.024]{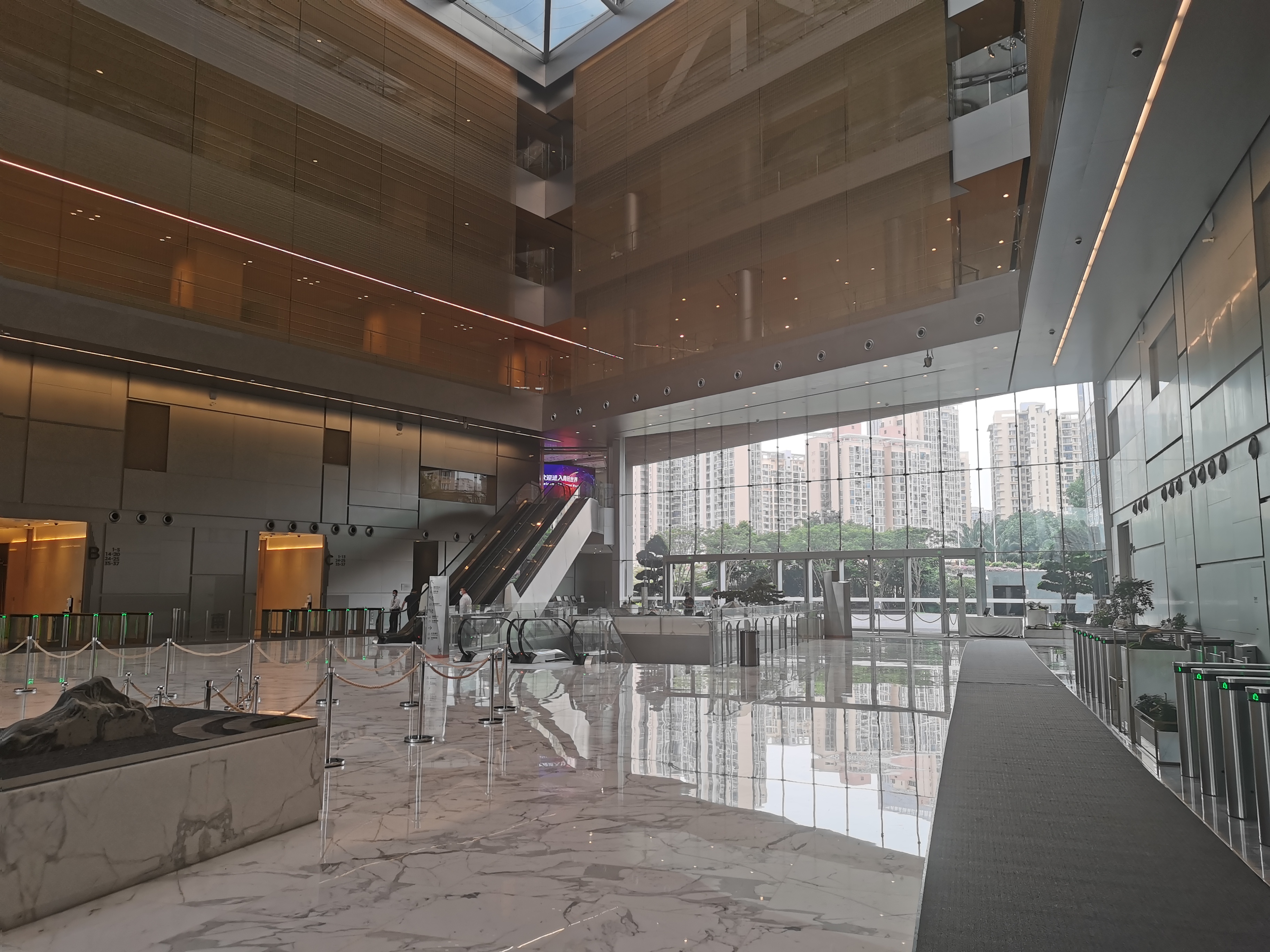}
				
				\\
				%(a) dCOEA &(b) PPS-RM &(c) DEE-DMOEA\\
				(a) & (b) & (c) & (d)\\
				\includegraphics[scale=0.024]{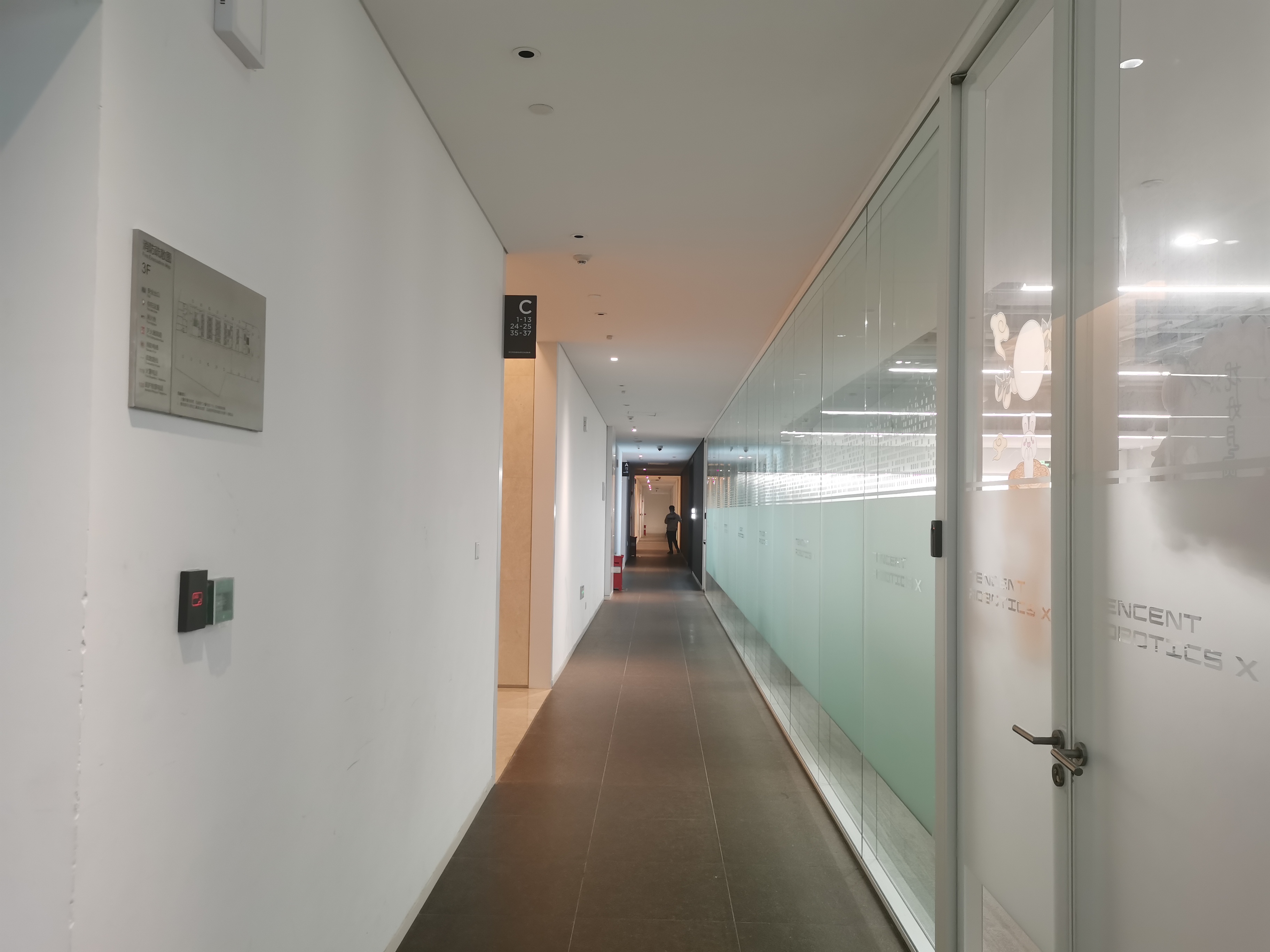} &
				\includegraphics[scale=0.024]{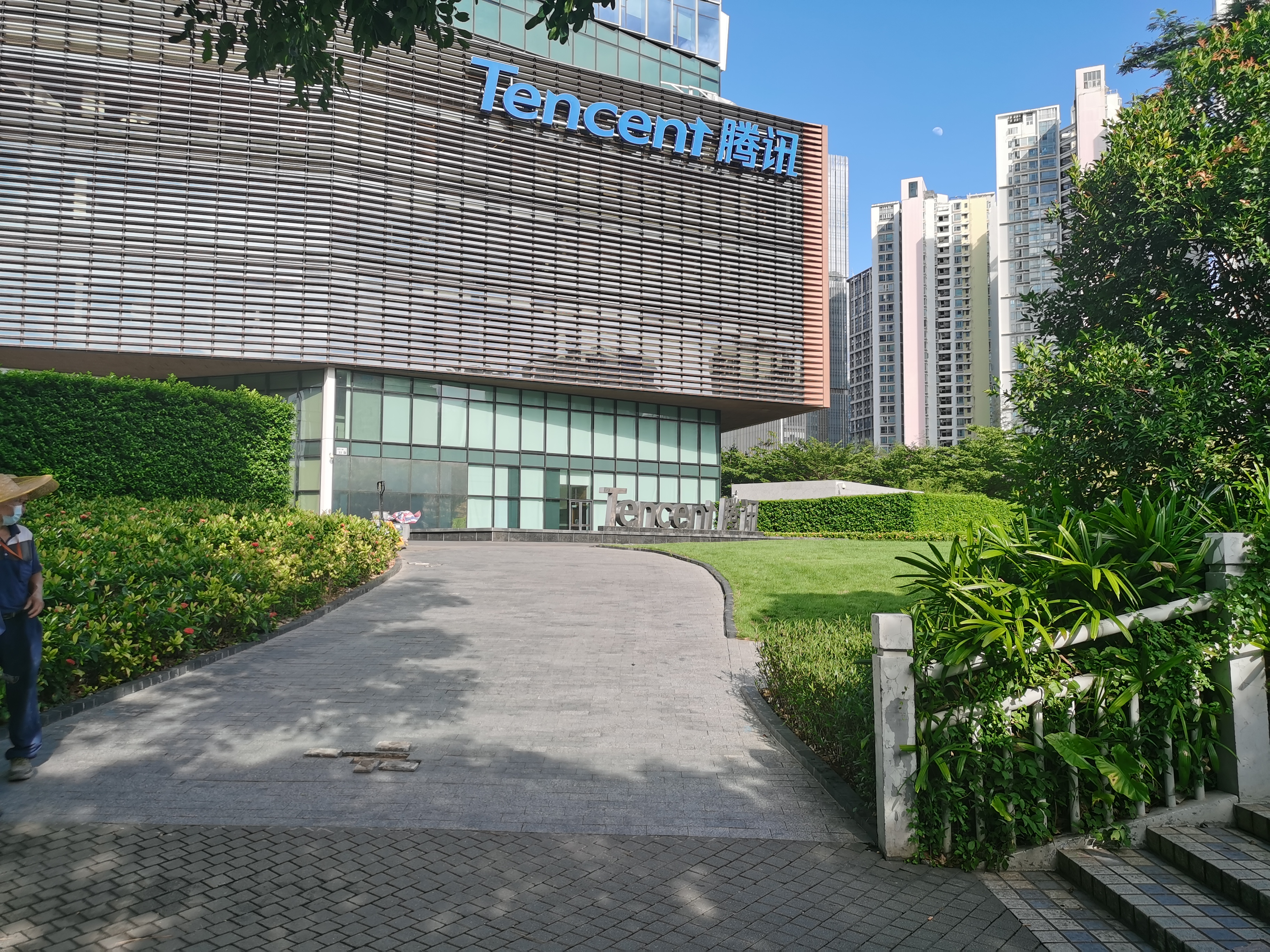} &
				\includegraphics[scale=0.024]{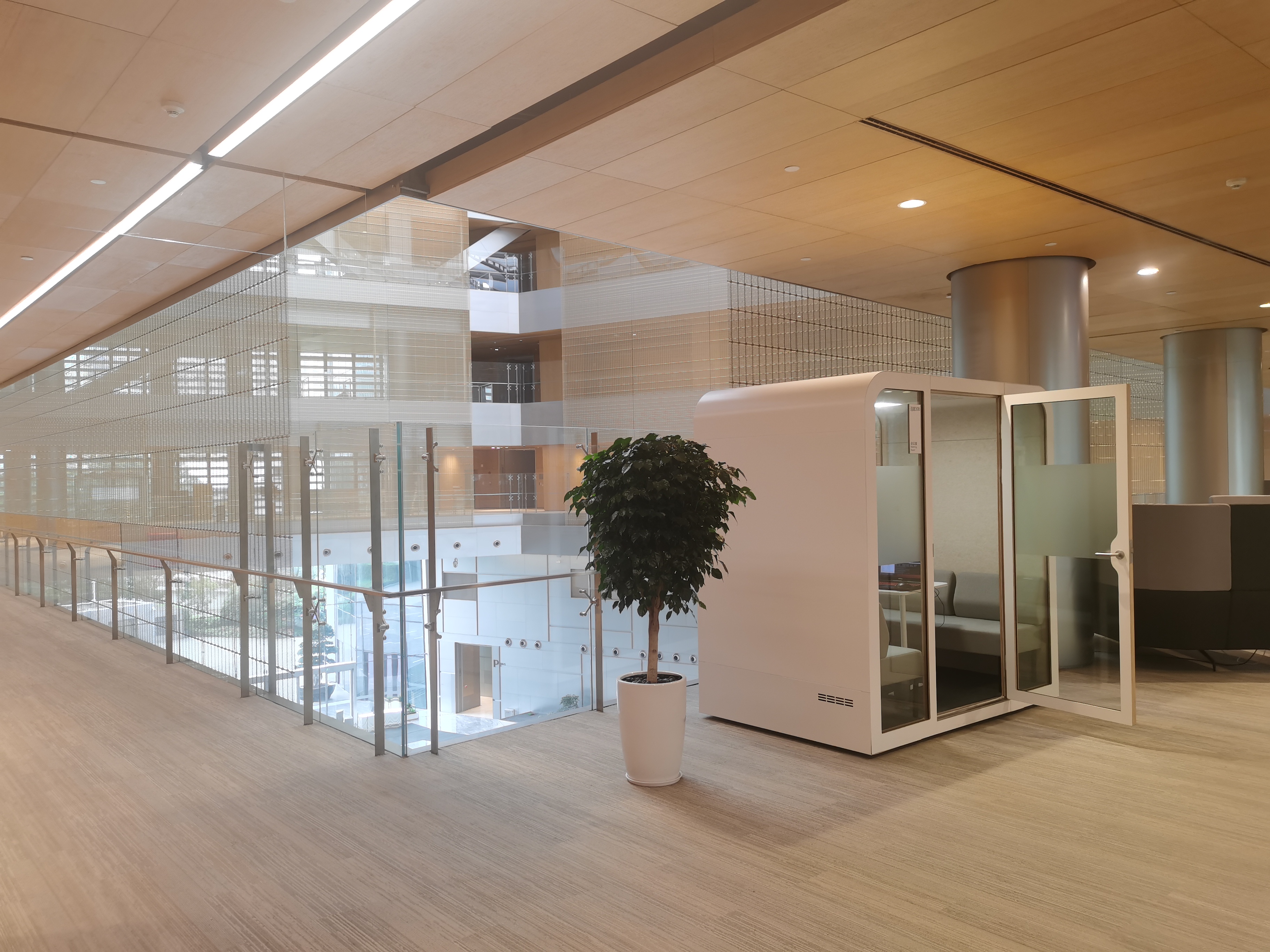}&
				\includegraphics[scale=0.024]{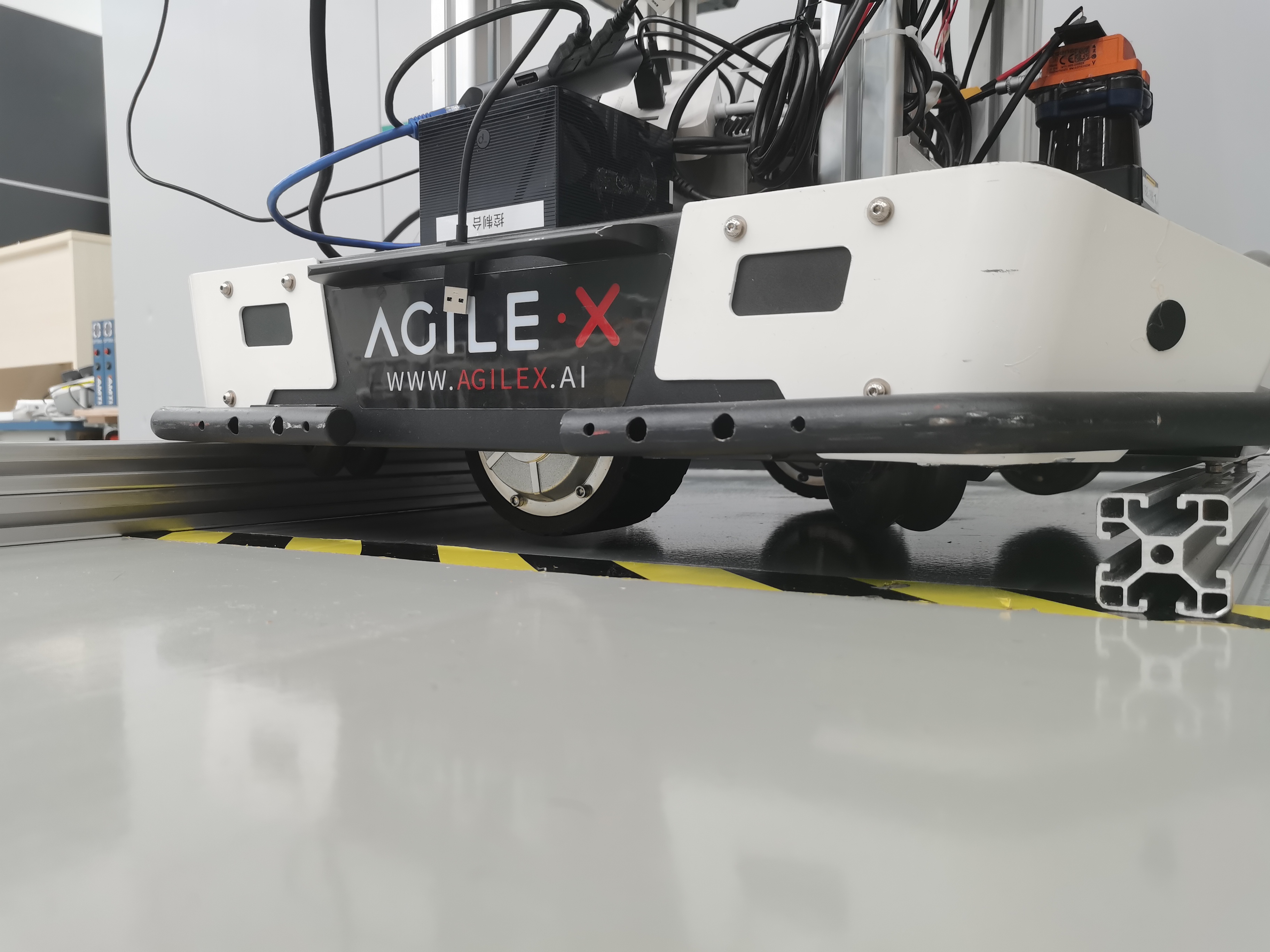}
				
				\\
				%(a) dCOEA &(b) PPS-RM &(c) DEE-DMOEA\\
				(e) & (f) & (g) & (h)\\			
			\end{tabular}
		\end{center}
		\caption{Diverse scenarios included in our datasets: (a). A room under the motion capture system. (b). a richly-textured and well-lit office. (c). A wall that lacks texture. (d). A hall with smooth floors. (e). A narrow corridor. (f). An outdoor slope. (g). Aisles with carpet. (h). Hanging the robot on a bracket. }
		\label{scenarios}
	\end{figure*}

\section{Introduction}

\IEEEPARstart{I}{ntelligent} ground robots have been widely used in industrial production and daily life, such as logistics, cleaning, warehouses, security, and food delivery. And navigation is the fundamental capability for these robots to execute these diverse tasks. To achieve reliable navigation, visual SLAM (Simultaneous Localization and Mapping) problem has been researched for decades, with quite a few classical methods proposed \cite{cadena2016past}. 
% Generally, there are two main types of SLAM systems: vision-based \cite{campos2021orb} and LiDAR-based methods \cite{qin2020lins} \cite{shan2018lego}. Although the LiDAR-based ones usually have better accuracy and robustness, the high cost of LiDARs makes them not available in some applications. 
% Therefore, vision-based SLAM algorithms have been a hot research issue in academia and industry.

A recent developing trend in visual SLAM is low-cost multi-sensor fusion, which has been verified to be a practical approach \cite{yin2021m2dgr}
to enhance the robustness to diverse scenarios. Different sensors can complement each other, maximizing the perceptual awareness of environments. One of the best example is that visual-inertial odometry (VIO) algorithms can significantly improve the tracking stability and accuracy in aggressive motion and textureless scenarios. 
While VIO systems have performed well in most cases, \cite{hesch2013consistency} has proven that this does not apply to ground vehicles.
For generic movement patterns, a VIO system has only four unobservable directions (three for global translation and one for global yaw). However, ground vehicles are restricted from moving in a 2D plane, mostly along a straight line or a circular arc, and thus the IMU is not sufficiently activated. 
Therefore, the VIO system on the ground robot will suffer from additional DoF unobservability, such as the scale. To address this issue, \cite{wu2017vins} extends VINS-Mono \cite{qin2018vins} to
incorporate low-frequency wheel-encoder data and keep the scale observable. Similarly, \cite{yang2019dre} proposes a RGB-D Encoder SLAM system for differential-drive robots. Most recently, \cite{Tingda2022VIW} proposes an optimization-based visual-inertial-wheel tightly coupled odometry, which claims to work robustly in dark or overexposed conditions. Nonetheless, its performance has not been tested on any public dataset with ground truth trajectories.

We believe that progress in SLAM, like in the AI field, is highly data-driven \cite{deng2009imagenet}. 
Although there have been extensive public datasets available to evaluate different SLAM algorithms, most of these datasets are outdated and do not challenge cutting-edge SLAM algorithms. In our opinion, those datasets focusing on challenging cases can more efficiently reveal the defects and limitations of existing algorithms. We notice that corner case detection in autonomous driving receive extensive concern from researchers \cite{bolte2019towards} \cite{muhammad2020deep} because such cases could easily cause the navigation system to drift. Similarly, once the localization module of the robot fails, it might cause industrial accidents and even pose potential threats to human safety as well. Nonetheless, to our knowledge, there is currently not much literature discussing the corner cases of robot navigation, which is not conducive to the safety of real-world robot applications.

To fill this gap, we present a novel SLAM dataset for ground robots, which aims to challenge existing cutting-edge SLAM systems with corner cases and thus promotes the progress of the multi-sensor fusion SLAM algorithm.
The challenges of our datasets lie in two areas: specific movement patterns and sensor failures, which will be elaborated in subsequent sections. Some scenarios covered in our datasets are visualized in Figure \ref{scenarios}. Our major contributions are summarized as follows:

\begin{itemize}
	\item We collect a novel visual SLAM dataset for ground robots with a rich pool of sensors in diverse environments both indoors and outdoors. Particularly, the dataset covers a series of challenging sequences including sensor failures and specific movement patterns.
	\item  State-of-the-art SLAM algorithms of different settings are tested on our benchmark. And the results indicate these systems are not robust enough for situations such as sensor failures.
	\item To facilitate the research on corner cases of robot navigation, we will release the full dataset with ground truth trajectories and the configuration file of each tested algorithm upon paper publication.
\end{itemize}

% The remaining paper is organized as follows: In Section II, the characteristics and defects of existing SLAM datasets are summarized. Section III details the sensor configuration and calibration, as well as the process of data collection. In Section IV, we evaluate the performance of state-of-the-art SLAM systems with different settings, and discuss the experiment results. Finally, Section V concludes this paper and outlines future work.

\section{Related Works}

\subsection{SLAM Datasets for Ground Robots}

% It is crucial to explore particular scenarios that fails localization and make targeted improvements to enhance the robustness of algorithms in these cases
% in consideration of safety. And dataset provides support for this.
	
Most existing SLAM datasets are collected by UAVs \cite{burri2016euroc} or cars \cite{ligocki2020brno}, but only a few are targeted at ground robots. For instance, Rawseeds \cite{bonarini2006rawseeds} and UTIAS\cite{leung2011utias} provide RGB images only, thus making them unsuitable for evaluating multi-sensor fusion systems. The Rosario dataset \cite{doi:10.1177/0278364919841437} is rich in sensor variety, yet is specifically designed for agricultural environments. M2DGR \cite{yin2021m2dgr} captures diverse indoor and outdoor scenarios, including some challenging scenes like elevators and darkrooms, but doesn't contain wheel odometer information which is essential for multi-sensor fusion SLAM algorithms due to its low cost and high precision. OpenLORIS\cite{shi2020we} offers rich sensor types in visual challenging scenarios such as highly dynamic markets and poorly exposed corridors, but wheel challenges or motion challenges are not included.

\begin{table}\scriptsize
		\caption{Comparison of SLAM datasets on ground robots}
		\label{dataset comparison}
		\centering
		\begin{tabular}{cccccc}
			\hline

			\makecell[c]{Dataset} & \makecell[c]{Environment} & \makecell[c]{RGB } & \makecell[c]{Depth} & \makecell[c]{IMU}  & \makecell[c]{Odom} \\
			\hline
			
			\makecell[c]{Rawseeds \cite{bonarini2006rawseeds}}  & \makecell[c]{In/Outdoors}   & \makecell[c]{\Checkmark}    & \makecell[c]{} & \makecell[c]{} &  \makecell[c]{ } \\ 	

			\makecell[c]{UTIAS\cite{leung2011utias}}  & \makecell[c]{Indoors}   & \makecell[c]{\Checkmark}    & \makecell[c]{} & \makecell[c]{} &  \makecell[c]{ } \\

			\makecell[c]{TUM RGBD \cite{sturm2012benchmark}}  & \makecell[c]{Indoors}   & \makecell[c]{\Checkmark}    & \makecell[c]{\Checkmark} & \makecell[c]{\Checkmark} &  \makecell[c]{ } \\ 
			
			\makecell[c]{NCLT \cite{carlevaris2016university}}  & \makecell[c]{In/Outdoors}   & \makecell[c]{\Checkmark}    & \makecell[c]{} & \makecell[c]{\Checkmark} &  \makecell[c]{ } \\ 
			
			\makecell[c]{Rosario\cite{doi:10.1177/0278364919841437	}}  & \makecell[c]{Outdoors}   &  \makecell[c]{\Checkmark}    & \makecell[c]{} & \makecell[c]{\Checkmark} &  \makecell[c]{\Checkmark} \\ 

% 			\makecell[c]{Newer College\cite{ramezani2020newer}}  & \makecell[c]{In/Outdoors}   &  \makecell[c]{\Checkmark}    & \makecell[c]{\Checkmark} & \makecell[c]{\Checkmark} &  \makecell[c]{\Checkmark} \\ 

			\makecell[c]{OpenLORIS \cite{shi2020we}}  & \makecell[c]{Indoors}   &  \makecell[c]{\Checkmark}    & \makecell[c]{\Checkmark} & \makecell[c]{\Checkmark} &  \makecell[c]{\Checkmark} \\ 
			
			\makecell[c]{SubT-Tunnel \cite{rogers2020test}}  & \makecell[c]{Outdoors}   &  \makecell[c]{\Checkmark}    & \makecell[c]{} & \makecell[c]{\Checkmark} &  \makecell[c]{} \\ 
% 			\makecell[c]{OBVision \cite{li2022odombeyondvision}}  & \makecell[c]{Outdoors}   &  \makecell[c]{\Checkmark}    & \makecell[c]{} & \makecell[c]{\Checkmark} &  \makecell[c]{} \\ 

			\makecell[c]{M2DGR\cite{yin2021m2dgr}}  & \makecell[c]{In/Outdoors}   & \makecell[c]{\Checkmark}    & \makecell[c]{} & \makecell[c]{\Checkmark} &  \makecell[c]{ } \\

			\hline
			\makecell[c]{Ours}  & \makecell[c]{In/Outdoors}   &  \makecell[c]{\Checkmark}    & \makecell[c]{\Checkmark} & \makecell[c]{\Checkmark} &  \makecell[c]{\Checkmark } \\
			\hline     
% 			\multicolumn{6}{l}{$^{\mathrm{a}}$We identify a dataset as long-term if it has sequences longer than 20 minutes.}
		\end{tabular}
	\end{table}

\subsection{Corner Cases}

Corner cases, i.e., extreme and non-predictable situations, are a popular research topic in autonomous driving \cite{bogdoll2021description}. Although infrequent, these cases can potentially threaten the security and reliability of autonomous navigation systems. Corner cases exist in robot navigation tasks as well. To address such challenging scenarios, researchers have proposed various methods, such as RGB-D SLAM \cite{liu2022rgb} and DS-SLAM \cite{yu2018ds}, to handle dynamic environments, and GVINS \cite{cao2022gvins} to deal with degenerate cases including low-speed movement, less than four visible satellites, and GNSS-denial environments. Additionally, \cite{zheng2022fast} proves that their method is robust in aggressive motions and a visual texture-less white wall. Nonetheless, we note that there are still plenty of corner cases that tend to be overlooked, such as wheel slippage, motion blur, and complete visual occlusion. There is a lack of SLAM datasets specifically designed for studying these corner cases, which is a gap yet to be filled. To sum up, it is urgent and critical to collect a novel SLAM dataset with rich sensor types, precise calibration, and sufficient challenge to support studies on corner cases, particularly sensor failures.

	\begin{table*}[h]
		\centering
		\caption{Specifications of sensors}
		\label{sensor}
		\begin{tabular}{cccc}
			\hline
			Device & Type  & Spec. & Freq.(Hz) \\
			\hline
			\multirow{3}[0]{*}{VI-sensor} & \multirow{3}[0]{*}{Realsense D435I} & RGB: 640*480, 69 H-FOV, 42.5V-FOV & 15 \\
			&       & Depth: 640*480, 0.1$\sim$10 meters & 15 \\
			&       & IMU: 6-axis & 200 \\

			IMU   & Xsens Mti-300 & 9-axis & 400 \\
			Wheel Odometer   & AgileX & 2D & 25 \\
			\hline
			LiDAR & Velodyne VLP-16 & 16 beam,360 H-FOV,30V-FOV & 10 \\

			\hline
		\end{tabular}%
		\label{tab:addlabel}%
	\end{table*}%

	\section{THE Ground-Challenge DATASET}
% 	\subsection{Acquisition platform}	
% 	We construct a ground robot for data collection as shown in Figure \ref{car}. 
	
% 	This robot has three layers. The bottom layer contains the power supply, the computer, and the display. The middle layer and the top layer include different sensors. The dimension figures of our robot are shown in Figure \ref{car} (a) $\sim$ (c).
% 	To ensure the high-speed data transmission, we connect the LiDAR to the Ethernet port of the host and other sensory devices to the USB3.0 port of the host. We record the data on a high-end laptop with a high-speed NMVe SSD.

	\begin{figure}
		\centering
		\includegraphics[scale=0.42]{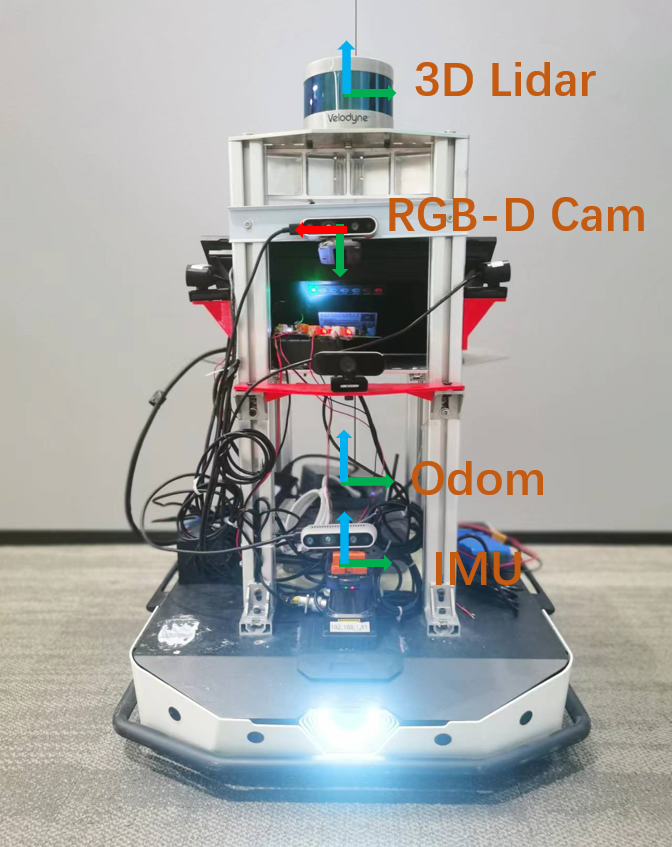}
		\caption{Our ground robot for data collection. Red is x-axis, green is y-axis, and blue is z-axis.   }
		\label{car}
	\end{figure}

	\subsection{Sensor setup}
We construct a ground robot for data collection and the sensor locations on the robot are shown in Figure \ref{car}. The chassis is equipped with a front-view VI-Sensor (Visual-Inertial Sensor) that captures RGB and depth images along with 6-axis IMU's measurements. Driven by two driving wheels providing odometer information and four assisting wheels, the robot also has a high-precision 9-axis Xsens IMU and a 16-beam 3D LiDAR.

The ground truth trajectories and point clouds are generated by the Velodyne LiDAR and the Xsens IMU using Fast-LIO2 \cite{xu2022fast}, a state-of-the-art LiDAR-based SLAM system. To evaluate its performance, we compared the high-precision trajectories generated by a motion capture system with 16 infrared cameras to those generated by Fast-Lio2. The experiment revealed that Fast-LIO2 can reach a positioning accuracy of 3cm in a small-scale (15m x 15m) indoor room. Additionally, as reported in \cite{xu2022fast}, Fast-LIO2 can achieve less than 0.1m end-to-end error in an outdoor trajectory spanning 1000 meters. Thus, considering that it is difficult for visually-based SLAM algorithms to achieve similar accuracy in challenging scenarios, we use the result of Fast-LIO2 as the pseudo-ground-truth trajectory.
	
	\begin{center}
	\begin{table}[h]
		\caption{An overview of scenarios in our dataset.}
		\label{overview}
		\centering
		\begin{tabular}{ccccc}
		%{p{1.15cm}p{1.15cm}p{1cm}p{1.05cm}p{1cm}}
			\hline

			{Scenario} & {Seq. Num} & {Dist./m}  & {Vel./(m/s)}& {Size/GB} \\
			\hline
			
			{Office} & {3} & {75.5}  & {0.46}& {4.7}  \\
			{Room} & {3} & {102.1}  & {0.66}& {4.6}   \\
			\hline		
			
			{Darkroom} & {3} & {92.0}  & {0.45}& {6.1}  \\
			
			{Wall} & {3} & {86.7}  & {0.46}& {5.6}   \\
			{Motionblur} & {3} & {166.6}  & {1.15}& {4.3}   \\
			
			{Occlusion} & {4} & {273.8}  & {0.82}& {9.9}  \\
			\hline

			{Roughroad} & {3} & {68.1}  & {0.37}& {5.4}   \\
			{Slope} & {2} & {128.5}  & {0.66}& {5.7}  \\

			{Hall} & {3} & {263.3}  & {0.87}& {8.7}  \\

			{Loop} & {2} & {371.8}  & {1.12}& {9.9}  \\
			
			\hline

			{Corridor} & {2} & {68.1}  & {0.37}& {5.4}\\

			{Rotation} & {3} & {12.4}  & {0.06}& {5.4}   \\

			{Static} & {2} & {1.9}  & {0.00}& {2.7}  \\

			\hline

			{Total} & {36} & {1780.0}  & {---}& {78.8}   \\

			\hline     
			
		\end{tabular}
	\end{table}
		\end{center}

% 	\begin{table}[h]
% 		\caption{Sample sequences for evaluation}
% 		\label{sequence feature}
% 		\centering
% 		\begin{tabular}{p{1.2cm}p{1.2cm}p{1cm}p{1.1cm}p{1.1cm}}
% 			\hline
% 			{Sequence} & {Duration/s} & {Dist/m}  & {Speed/(m/s)}  & {Size/GB}  \\
% 			\hline

% 			{Office3} & {52.2}  & {16.7}  & {0.32}  & {1.5}  \\
			
% 			\hline
			
% 			{Darkroom2} & {76.5}  & {38.1}  & {0.50}  & {2.3}  \\
			
% 			{Wall2} & {66.0}  & {26.1}  & {0.40}  & {2.0}  \\	

% 			{Motionblur3} & {39.5}  & {41.0}  & {1.05}  & {1.2}  \\
% 			{Occlusion4} & {39.9}  & {28.8}  & {0.72}  & {1.2}  \\
% 			\hline

% 			{Roughroad3} & {58.9}  & {17.9}  & {0.30}  & {1.8}  \\
			
% 			{Slope1} & {95.8}  & {40.2}  & {2.38}  & {2.8}  \\
			
% 			{Hall1} & {91.0}  & {77.8}  & {0.85}  & {2.6}  \\
% 			{Loop2} & {195.7}  & {208.5}  & {1.07}  & {5.8}  \\

% 			\hline
			
% 			{Corridor1} & {100.3}  & {79.9}  & {0.80}  & {2.9}  \\
	
% 			{Rotation3} & {57.5}  & {5.2}  & {0.09}  & {1.7}  \\

% 			{Static1} & {55.8}  & {1.2}  & {0.02}  & {1.6}  \\

% 			\hline     
			
% 		\end{tabular}
% 	\end{table}

	\begin{table*}[h]\scriptsize
		\caption{ ATE RMSE (m) of SLAM systems on sample sequences}
		\label{ate rmse tab}
		\centering
		\begin{tabular}{cccccccc}
		%{p{2.5cm}p{2.2cm}p{2.2cm}p{2cm}p{2cm}p{1.8cm}p{2cm}}
			\hline
			{ Type} &{ Sequence / Method} & {VINS-Mono \cite{qin2018vins}} & {VINS-RGBD \cite{shan2019rgbd}}  & {VIW-Fusion \cite{Tingda2022VIW}}  & {EKF \cite{ribeiro2004kalman}} & {Raw Odometer}\\
			\hline
			
			{Normal}&{Office3}&  {0.35}  & {0.31}  & {0.18}  & {0.16} & {\textbf{0.16}}  \\
			\hline

			{}&{Darkroom2} & {1.66}  & {0.82}  & {0.53}  & {\textbf{0.22}} & {0.30}  \\
		
% 			{Occlusion2}& {---} & {3.62}  & {4.16}  & {0.31}  & {9.56} & {0.85}  \\

			{Visual}&{Wall2} & {1.21}  & {1.00}  & {\textbf{0.15}}  & {32.91} & {0.57} 
			\\
			
			{Challenge}&{Motionblur3} & {9.37}  & {32.31}  & {0.78}  & {\textbf{0.42}} & {0.79}  \\	
			{}&{Occlusion4}& {*$^{\mathrm{a}}$}  & {*}  & {*}  & {\textbf{0.22}} & {0.23}  \\
						
			\hline
			{}&{Roughroad3}&  {0.17}  & {25.52}  & {0.14}  & {\textbf{0.11}} & {\textbf{0.11}}  \\

			{Wheel}&{Slope1}&  {9.41}  & {2.84}  & {\textbf{0.65}}  & {0.89} & {0.89}  \\

			{Challenge}&{Hall1}&  {7.06}  & {94.27}  & {\textbf{0.85}}  & {2.16} & {2.79}  \\

			{}&{Loop2}&  {6.09}  & {\textbf{3.44}}  & {9.23}  & {16.70} & {17.61}  \\
			\hline

			{Motion}&{Corridor1} &  {4.48}  & {\textbf{0.85}}  & {1.12}  & {1.78} & {2.17} \\

			{Challenge}&{Rotation3}&  {29.12}  & {0.19}  & {0.18}  & {\textbf{0.14}} & {\textbf{0.14}}  \\

			{}&{Static1}&  {*}  & {*}  & {*}  & {5.61} & {\textbf{3.54}}  \\

			\hline
			\multicolumn{7}{l}{$^{\mathrm{a}}$If a SLAM system fails to initialize or track frames less than a half of total frames, we mark it *.}
% 			\multicolumn{8}{l}{$^{\mathrm{a}}$If a visual SLAM fails to initialize or track frames less than a half of total frames or a GNSS-based method fails to initialize, we mark it X}
			
		\end{tabular}
	\end{table*}

	\subsection{Synchronization and Calibration}
We capture all the data using the ROSbag tool in the Robot Operating System (ROS). The RGB camera and 6-axis IMU embedded in the Realsense D435I are hard-synchronized, while the depth images are pixel-by-pixel aligned to the RGB images. The 3D LiDAR and 9-axis IMU are software-synchronized by triggering data capture at the same instance. To calculate the camera intrinsics of pinhole cameras, we use the MATLAB Camera Calibration Toolbox. To calibrate the internal parameters of the IMU, we use the toolbox from \cite{UCAM-CL-TR-696}, which includes the white noise and random walk of both the gyroscopic and accelerometer measurements. We choose the IMU frame as the reference to calibrate the extrinsic parameters (relative poses) between sensors, and employ the toolbox from \cite{furgale2013unified} for calibrating the extrinsic parameters between cameras and IMU.

	\subsection{Data collection}

We provide an overview of our dataset in Table \ref{overview}. All data was captured using the Rosbag tool within the Robot Operating System (ROS). The recording process is as follows: First, we recorded Office and Room sequences, where the robot moves slowly in a well-lit and textured office or room respectively, to test the performance of different algorithms in normal situations. Subsequently, we designed a series of corner case experiments from three aspects: visual challenge, wheel odometer challenge, and particular movement pattern, which are presented as follows:
	
		\begin{figure}
		\begin{center}
			\begin{tabular}{cc}
				\includegraphics[scale=0.11]{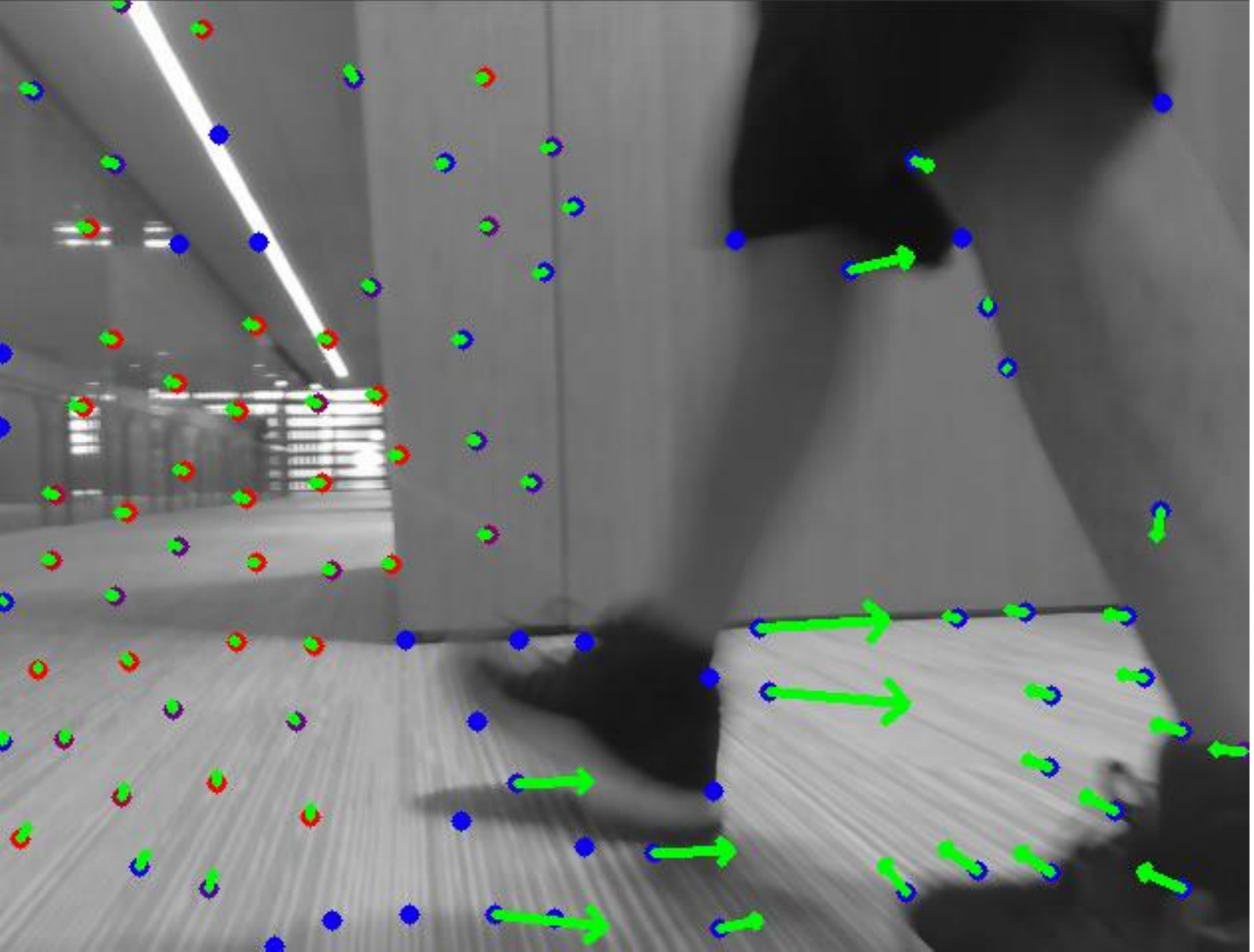} &
				\includegraphics[scale=0.11]{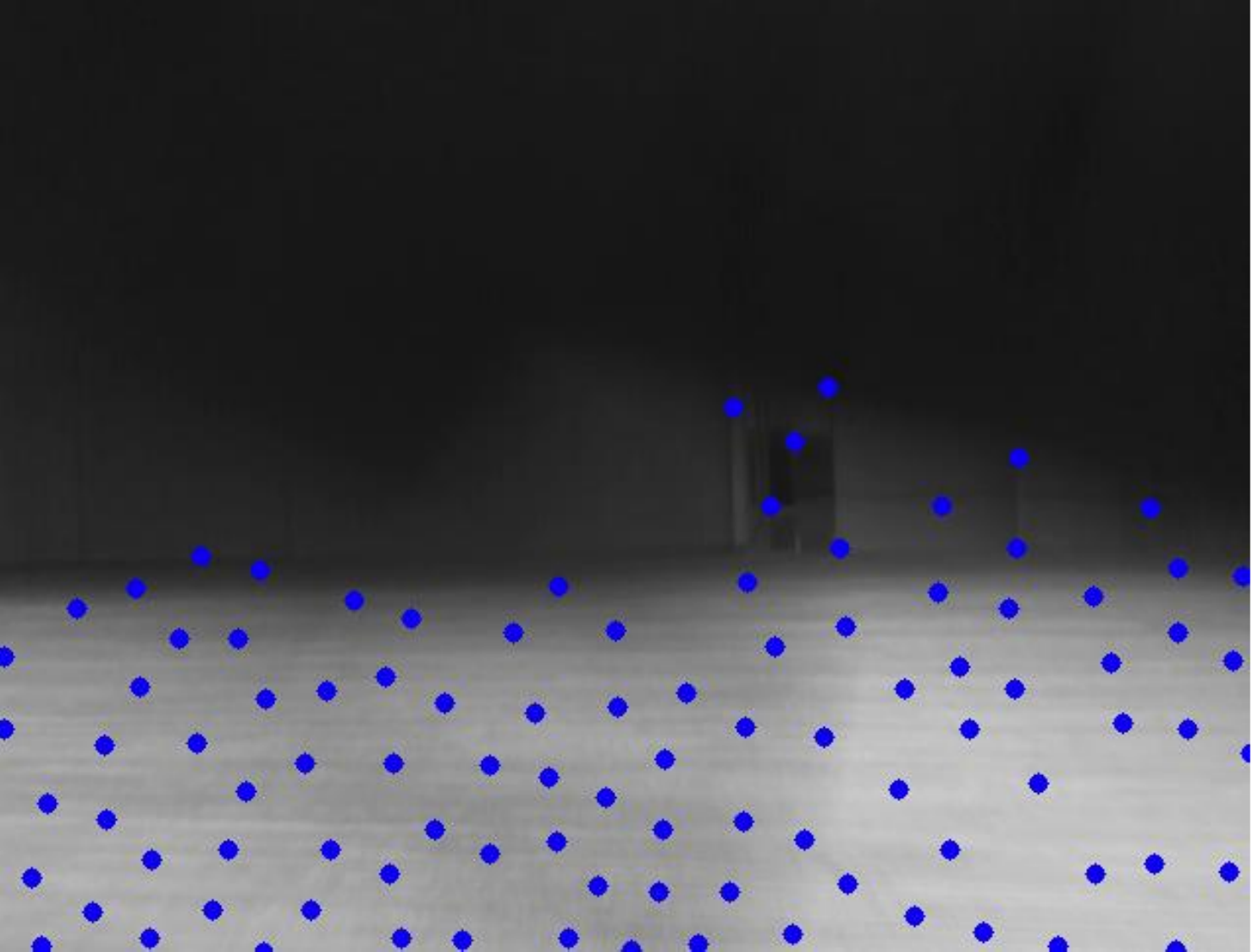} \\
				(a) & (b) \\
				\includegraphics[scale=0.11]{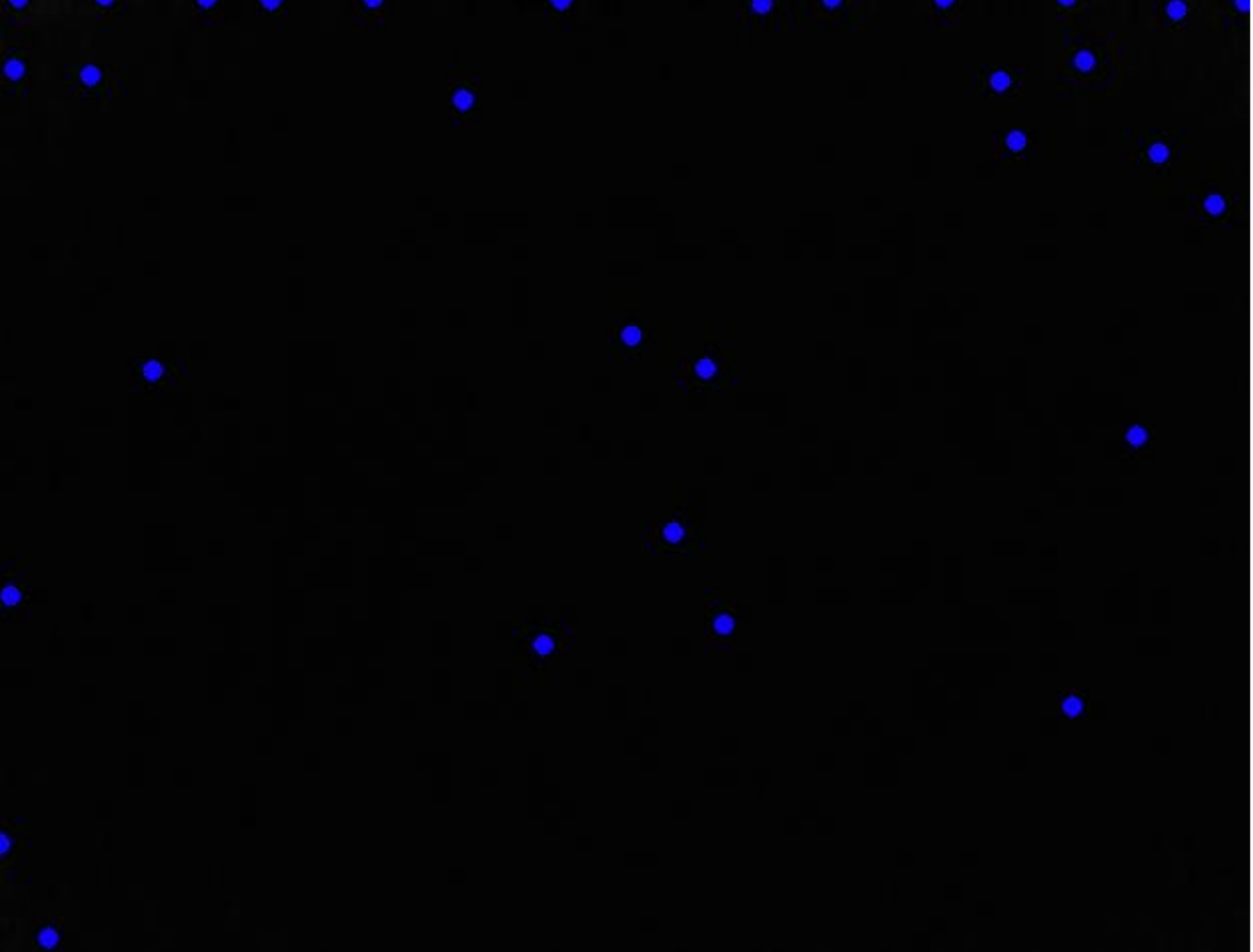}&
				\includegraphics[scale=0.11]{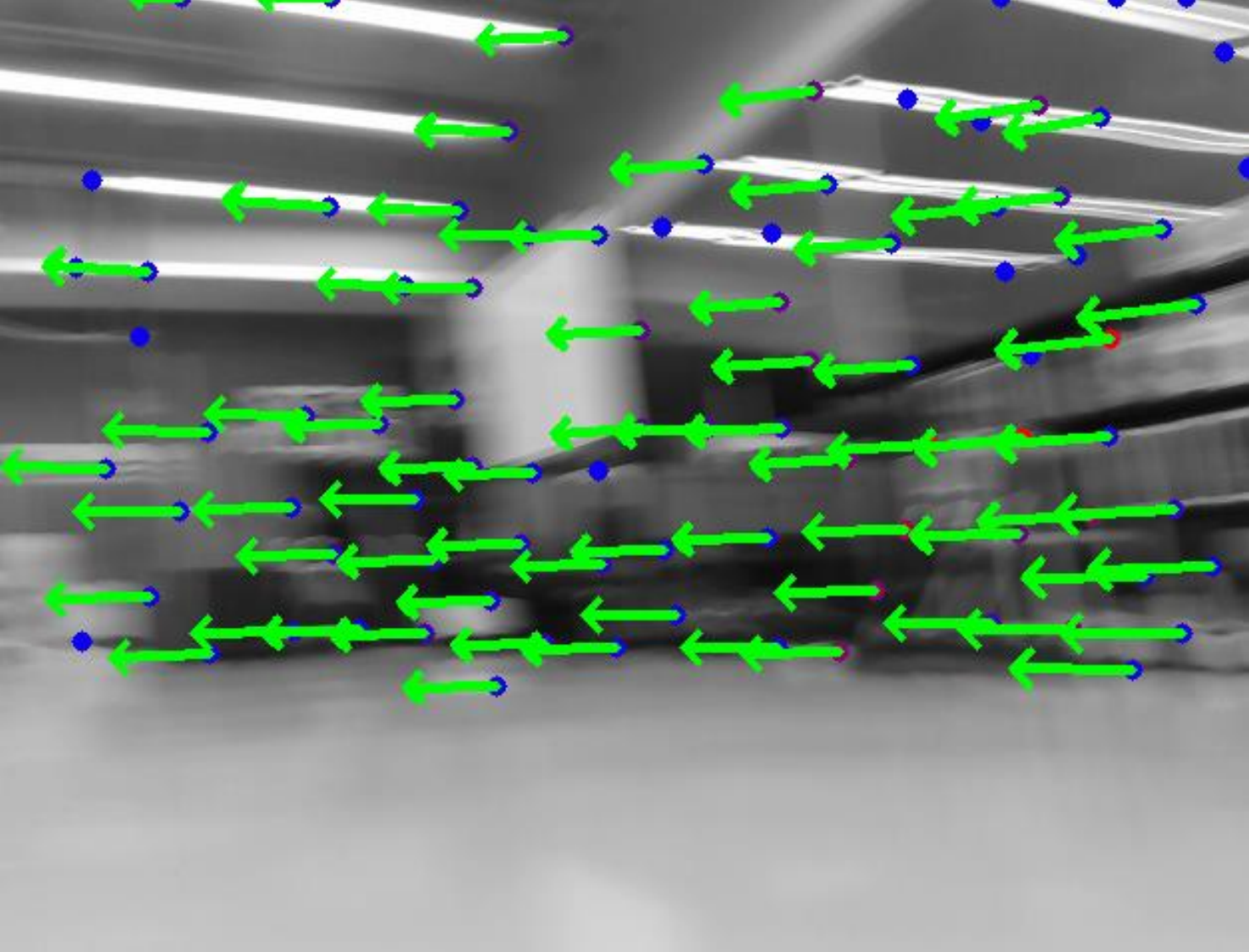}
				
				\\
				%(a) dCOEA &(b) PPS-RM &(c) DEE-DMOEA\\
				  (c) & (d)\\

			\end{tabular}
		\end{center}
		\caption{(a) Moving feet. (b) occluding the camera with a palm. (c) complete occlusion. (d) motion blur   }
		\label{occlusion}
	\end{figure}

	\subsubsection{Visual Challenge}
	In our experiments, we manipulate the robot to move in a room with poor illumination (Darkroom sequences), back and forth in front of walls lacking texture (Wall sequences), and through scenarios of varying degrees of occlusion (Occlusion sequences). Figure \ref{occlusion} (a) shows sequences Occlusion1$\sim$2, which involves a person walking in front of the robot and causing intermittent partial occlusion. Figure \ref{occlusion} (b) displays sequence Occlusion3, in which the camera is covered with the palm repeatedly. In sequence Occlusion4 (Figure \ref{occlusion} (c)), a piece of black tape is attached to the camera's lens to completely block its view, disabling feature extraction and matching for visual SLAM. Furthermore, Motionblur sequences are generated by rapidly translating and rotating the robot, creating motion blur for cameras (Figure \ref{occlusion} (d)).

	\subsubsection{Wheel Odometer Challenge}
	The Hall and Loop sequences are collected in a hall with smooth ground and a heavily carpeted aisle loop, respectively, where the wheels slip significantly. Moreover, we record Roughroad sequences to test the performance of the localization algorithm on rough roads.

	\subsubsection{Particular Moving Patterns}
	
    In the Sequences Corridor1 and Corridor2, the robot moves forward in a zigzag shape and straight forward, respectively. In the zigzag route, motion blur and less overlapping between adjacent image frames will lead to errors in feature matching. 
    In the Rotation sequence, the robot only rotates and hardly translates, which makes it difficult for vision-based algorithms to estimate the depth of feature points by triangulation. In the Static sequences, the robot stands still on a bracket, and we control its wheels to move in different directions through the handle. This experiment aims to test whether SLAM systems coupled with the wheel odometer can work well when the robot wheel is suspended.
	Finally, we operate the robot from a flat surface to another, passing through a slope. In this experiment, since the wheel odometer only provides two-dimensional speed observations, it could be misleading to estimate three-dimensional trajectories.

% To make the results easy to reproduce, we open source all the truth trajectories, calibration files, and configuration files of each tested algorithm on our project website.

	\begin{figure*}
		\begin{center}
			\begin{tabular}{cccc}
				\includegraphics[scale=0.26]{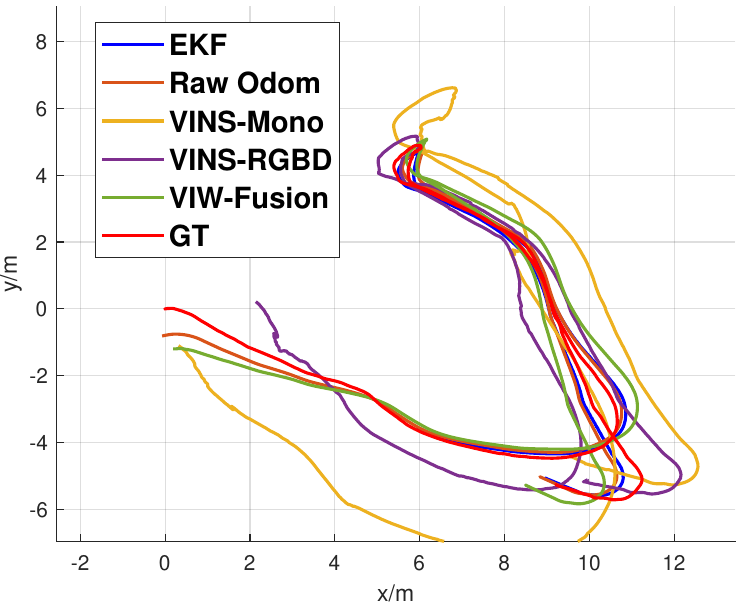} &
				\includegraphics[scale=0.26]{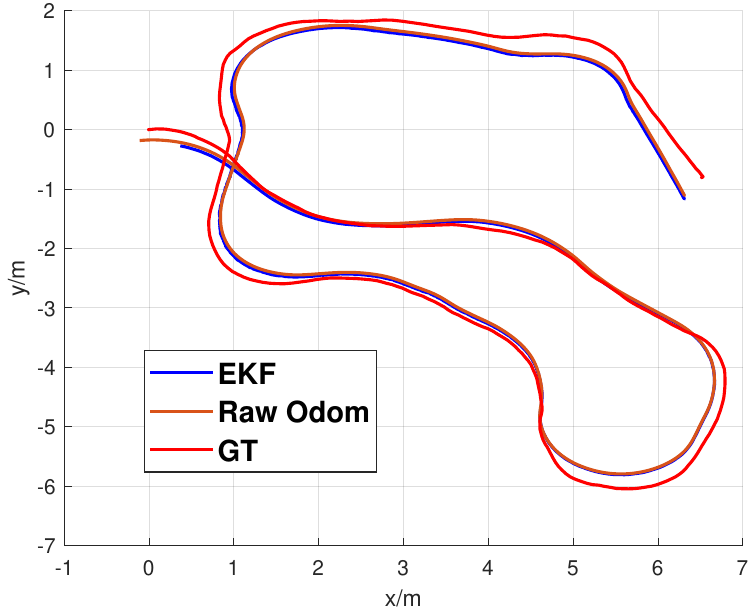} &
				\includegraphics[scale=0.26]{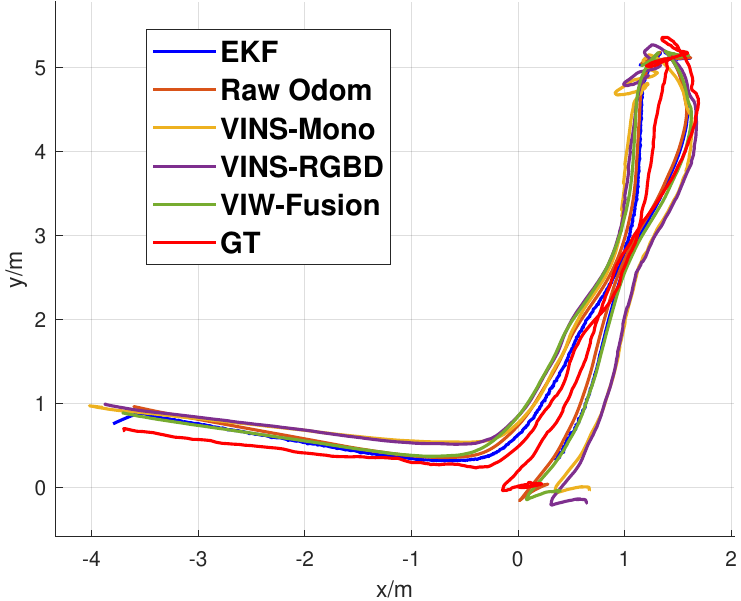}&
				\includegraphics[scale=0.26]{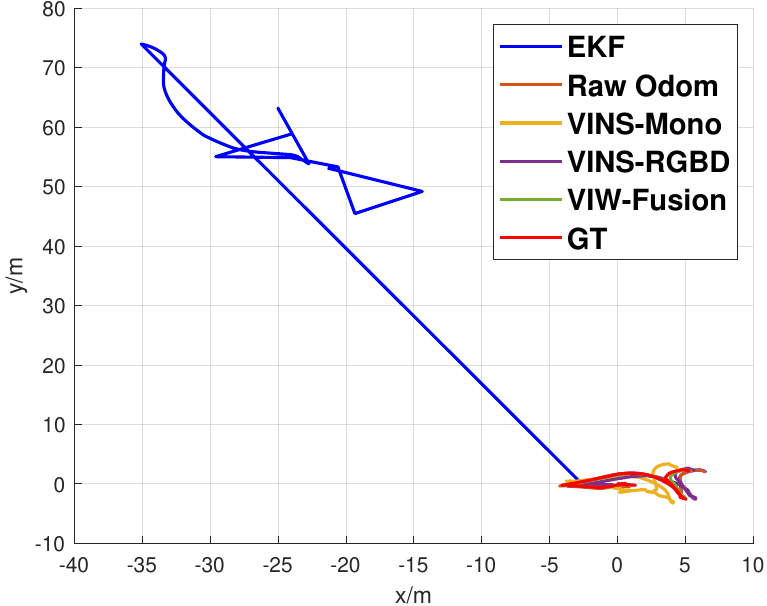}
				
				\\
				%(a) dCOEA &(b) PPS-RM &(c) DEE-DMOEA\\
				(a). Darkroom2 & (b). Occlusion4 & (c).  Office3 & (d).  Wall2\\
				\includegraphics[scale=0.26]{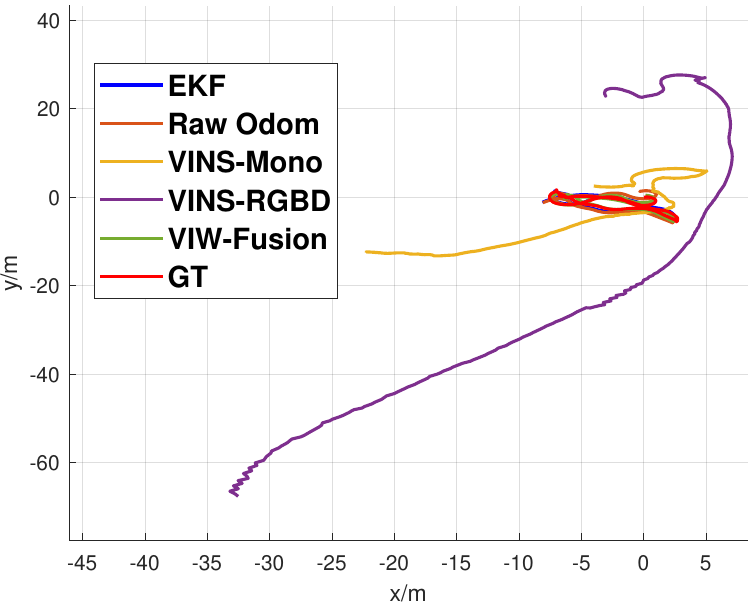} &
				\includegraphics[scale=0.26]{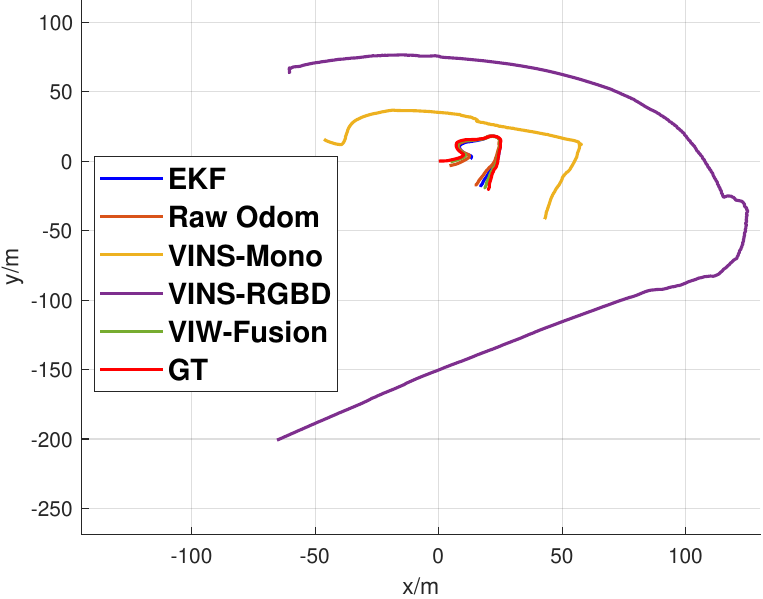} &
				\includegraphics[scale=0.24]{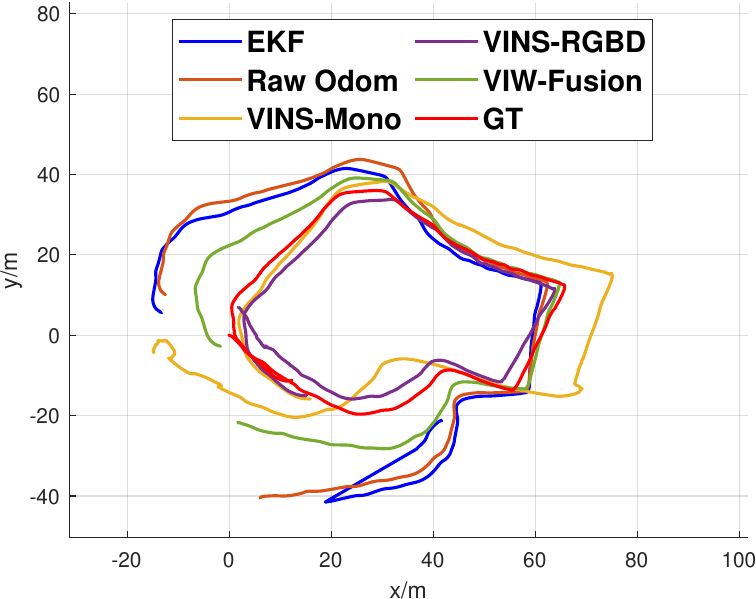}&
				\includegraphics[scale=0.26]{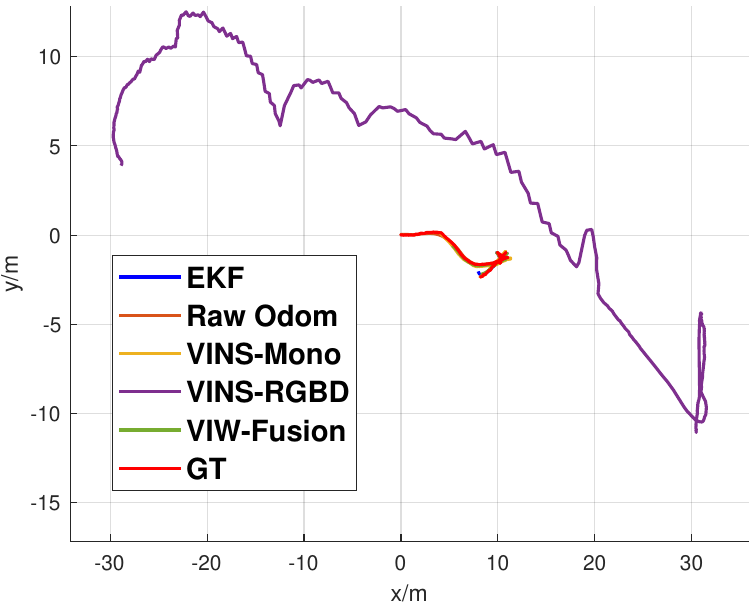}
				
				\\
				%(a) dCOEA &(b) PPS-RM &(c) DEE-DMOEA\\
				(e). Motionblur3 & (f).  Hall1 & (g).  Loop2 & (h).  Roughroad3\\

				\includegraphics[scale=0.26]{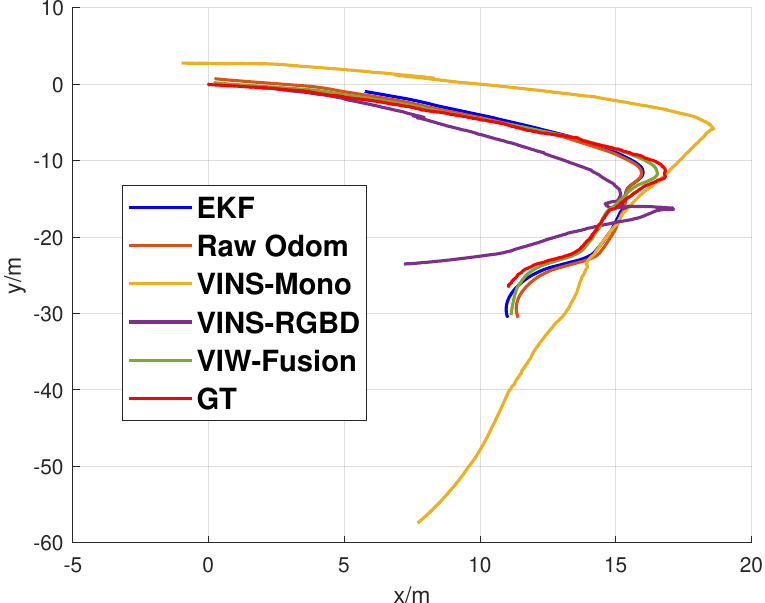} &
				\includegraphics[scale=0.26]{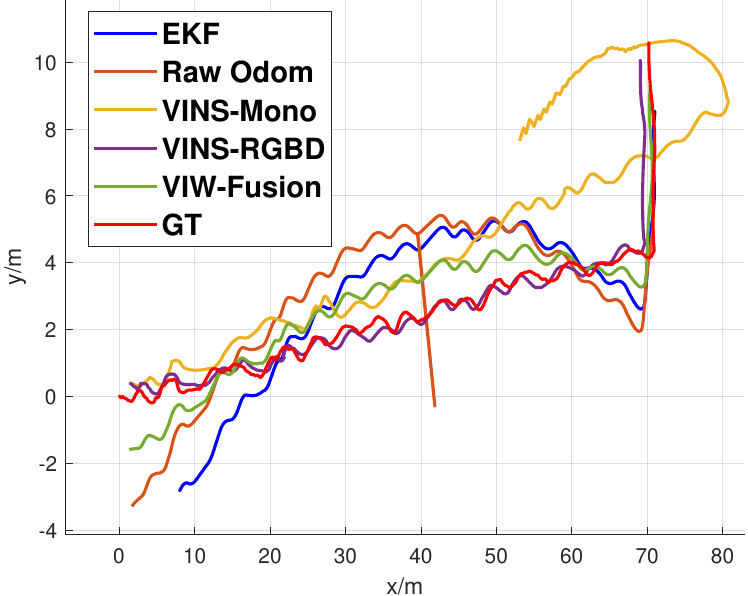} &
				\includegraphics[scale=0.26]{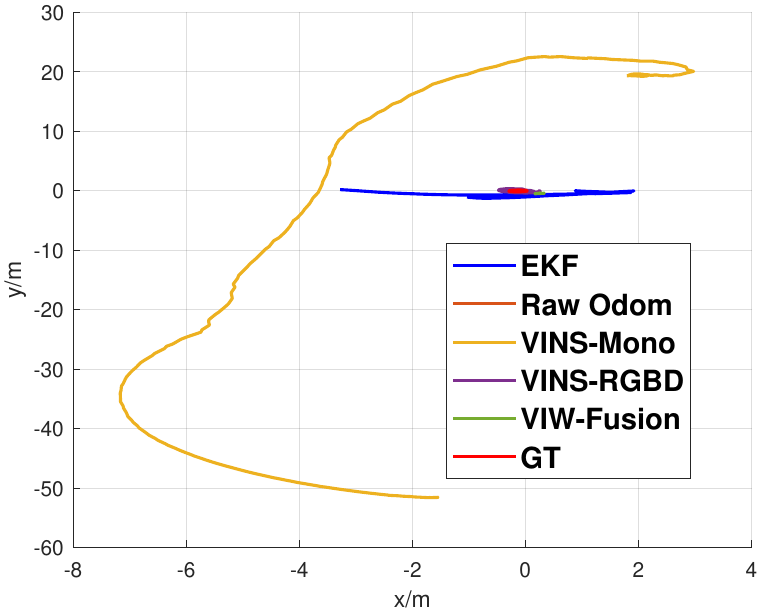}&
				\includegraphics[scale=0.26]{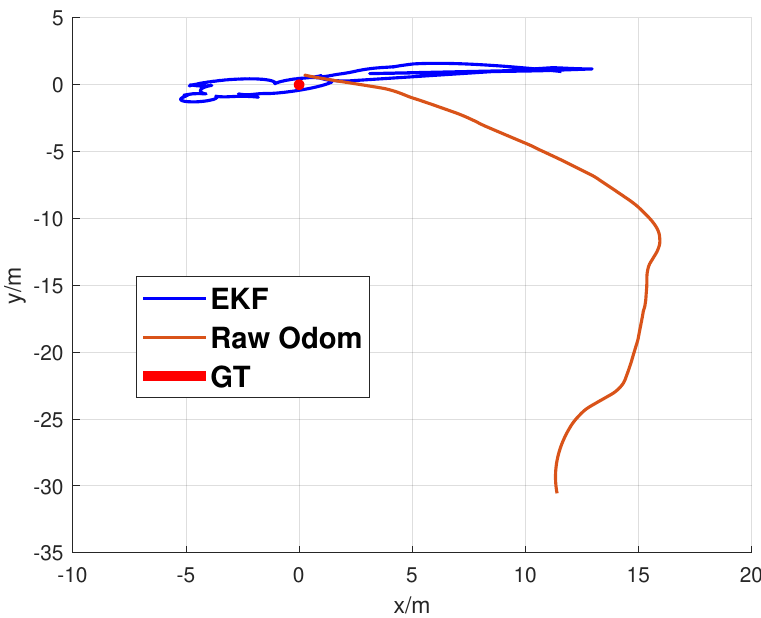}
				
				\\
				%(a) dCOEA &(b) PPS-RM &(c) DEE-DMOEA\\
				(i).  Slope1 & (j).  Corridor1 & (k).  Rotation3 & (l).  Static1\\	
			\end{tabular}
		\end{center}
		\caption{Estimated and ground-truth (GT) trajectories of 12 sample sequences are visualized on the x-y plane. }
		\label{plots}
	\end{figure*}

\section{Evaluation}
	
The features of all the sequences are described on our project website. We evaluated some SLAM systems with different sensor configurations on twelve representative sequences from our dataset. The tested algorithms are ORB-SLAM3 \cite{campos2021orb}, an optimization-based SLAM system; VINS-Mono \cite{qin2018vins}, one of the state-of-the-art monocular visual-inertial systems; VINS-RGBD \cite{shan2019rgbd}, a fusion algorithm of RGB-D and IMU information based on the VINS-Mono \cite{qin2018vins} framework; and VIW-Fusion \cite{Tingda2022VIW}, a tightly-coupled visual-inertial-wheel system featuring online extrinsic calibration and wheel-aided initialization. Also, we use an EKF algorithm \cite{ribeiro2004kalman} for fusion of IMU and wheel odometer.

The EVO tool \cite{MichaelGrupp2018} was used to align all the estimated trajectories with ground truth trajectories to obtain the ATE RMSE \cite{sturm2012benchmark}.
The quantitative results are shown in Table \ref{ate rmse tab}, with the estimated trajectories in 2D plotted in Figure \ref{plots}. Since most of the selected sequences are highly challenging (even with sharp turns), ORB-SLAM3 (both monocular-inertial and RGBD-inertial version) performed poorly on most of our test sequences, with frequent tracking failures (less than 50$\%$ of successfully tracked frames), initialization failure, or scale drift. 
In contrast, SLAM algorithms with multi-sensor fusion (like VIW-Fusion \cite{Tingda2022VIW}) achieved better localization results but failed in some specific scenarios as well. We discuss the experiment results in detail as follows:

\paragraph{Normal Situation} 
The ATE RMSE results on Sequence Office3 indicate that existing localization methods can perform well when the motion mode matches the assumptions of these algorithms and all the sensors work well.

\paragraph{Vision Challenge} 

In Sequence Darkroom2 and Motionblur3, VINS-Mono \cite{qin2018vins} and VINS-RGBD \cite{shan2019rgbd} drift a lot due to visual failures, while Wheel odometer based algorithms work more robustly in this case.
In Sequence Occlusion4, all the vision-based methods including VIW-Fusion \cite{Tingda2022VIW} fail to initialize because of poor feature extraction. This finding indicates that VIW-Fusion \cite{Tingda2022VIW} has not been adequately designed to handle adverse conditions. A more prudent strategy may be to combine the wheel odometer and IMU to output a trajectory when a visual sensor failure is detected.

	\paragraph{Wheel Odometer Challenge}

In the sequences Roughroad3 and Slope1, vision-based systems perform worse than wheel odometer-based algorithms due to inaccurate scale estimation in aggressive motion. In Sequence Hall1, VINS-Mono \cite{qin2018vins} and VINS-RGBD \cite{shan2019rgbd} drift significantly due to ground reflection and faraway feature points. Here, VIW-Fusion \cite{Tingda2022VIW} maintains satisfactory positioning performance even with slight wheel slippage, demonstrating the advantages and necessity of multi-sensor fusion in complex scenarios. However, when the wheels slip more severely in Sequence Loop2, the significant deviation caused by the wheel odometer increases the localization error of estimated trajectories. This can be attributed to two main reasons: current algorithms lack the ability to detect wheel slippage, and the angular velocity provided by the wheel speedometer is not accurate, leading to the long-term divergence of the estimated trajectory. To reduce the accumulation of errors, it is suggested that IMU's angular velocity measurement be used instead of the wheel odometer's.

% which reveals that current wheel odometer-based algorithms are under-considered for wheel slip situations.

\paragraph{Particular Movement Patterns}
In Sequence Corridor1, the zigzag movement of the robot not only fails the feature extraction but also leads to severe wheel slippage. Therefore, all the tested algorithms cannot accurately estimate the trajectory. In Sequence Rotation1, pure rotation causes severe errors in depth estimation by VINS-Mono's triangulation, while the remaining tested systems perform well thanks to measurements from other sensors. Finally, in Sequence Static1, VIO systems cannot be initialized successfully due to the lack of IMU excitation. Since the wheels are still moving after suspension, the wheel odometer-based methods mistake the robot being in motion.

In summary, VINS-Mono \cite{qin2018vins} is most likely to generate catastrophic localization results in corner cases, and VINS-RGBD \cite{shan2019rgbd} can also inevitably fail when severe camera failures occur. 
We have noticed that the wheel odometer alone can achieve good results in most situations, except for severe wheel slippage. Integrating the IMU and the wheel odometer through the EKF \cite{ribeiro2004kalman} can achieve higher accuracy than the raw odometer. Nonetheless, the trajectory of the EKF can shake violently in the initialization phase due to the inaccuracy in the initial covariance estimation (this part was manually eliminated in our experiment). VIW-Fusion \cite{Tingda2022VIW} can achieve satisfying accuracy and robustness in most sequences, but its initialization in visual failure needs improvement. Furthermore, it lacks consideration for wheel slippage, and its adopted dead reckoning model will diverge in a long trajectory due to inaccurate angular velocity estimates.

The experiments conducted demonstrate the validity and value of our dataset as a benchmark for existing SLAM systems. The results further suggest that there is still much room for improvement in current cutting-edge multi-sensor fusion algorithms for real-world applications. Sensor failures, such as complete occlusion and wheel suspension, can be fatal for single-sensor-based methods; however, multi-sensor fusion systems should be designed to be more robust in these cases. For instance, we posit that a reliable visual-IMU-wheel system should be able to explicitly identify scenarios where visual observations are inaccurate and respond accordingly (e.g. disable visual information and rely only on wheel odometer and IMU). Nevertheless, to our knowledge, corner case identification and troubleshooting have been scarcely addressed in prior work. Therefore, we provide this dataset to support relevant researches.

	\section{Conclusion}
We present Ground-Challenge, a novel ground robot dataset to encourage breakthroughs in multi-sensor fusion SLAM algorithms. Specifically, we have crafted a series of corner case experiments, including sensor failures in diverse environments, to challenge current cutting-edge SLAM systems. We have tested these systems on our dataset and analyzed their limitations in various scenarios, thus providing potential developing directions for SLAM. We are committed to continually updating our benchmark dataset. Specifically, we will mount 2D and 3D LiDAR on the robot, design experiments to invoke corner cases, and utilize higher-precision equipment such as motion capture systems to ensure accurate ground truth for LiDAR SLAM in our future work. 

\noindent
\textbf{Acknowledgement}
Thank Tencent Robotics X Lab for support to this work.

\ifCLASSOPTIONcaptionsoff
  \newpage
\fi

\bibliographystyle{IEEEtran}
\bibliography{main}

\end{document}